\documentclass[sigconf]{acmart}
\usepackage{multirow}
\usepackage[ruled,linesnumbered]{algorithm2e}
\usepackage{bm}
\AtBeginDocument{%
  \providecommand\BibTeX{{%
    \normalfont B\kern-0.5em{\scshape i\kern-0.25em b}\kern-0.8em\TeX}}}
    
\copyrightyear{2020}
\acmYear{2020}
\setcopyright{acmcopyright}\acmConference[GLSVLSI '20]{Proceedings of the Great Lakes Symposium on VLSI 2020}{September 7--9, 2020}{Virtual Event, China} \acmBooktitle{Proceedings of the Great Lakes Symposium on VLSI 2020 (GLSVLSI '20), September 7--9, 2020, Virtual Event, China}
\acmPrice{15.00}
\acmDOI{10.1145/3386263.3407650}
\acmISBN{978-1-4503-7944-1/20/09}

\begin{document}
\fancyhead{}
\title{A Privacy-Preserving-Oriented DNN Pruning and Mobile Acceleration Framework}


\author{Yifan Gong$^{1}$, Zheng Zhan$^{1}$, Zhengang Li$^{1}$, Wei Niu$^{2}$, Xiaolong Ma$^{1}$, Wenhao Wang$^{4}$, Bin Ren$^{2}$, Caiwen Ding$^{3}$, Xue Lin$^{1}$, Xiaolin Xu$^{4}$, and Yanzhi Wang$^{1}$}
\affiliation{%
  \institution{$^{1}$Northeastern University $^{2}$The College of William and Mary $^{3}$University of Connecticut $^{4}$University of Illinois at Chicago}}
\email{{gong.yifa, zhan.zhe, li.zhen, ma.xiaol, xue.lin, yanz.wang}@northeastern.edu}
\email{wniu@email.wm.edu, bren@cs.wm.edu, caiwen.ding@uconn.edu, {wwang208,xiaolin8}@uic.edu}

\renewcommand{\shortauthors}{Yifan Gong and Zheng Zhan, et al.}

\begin{abstract}
Weight pruning of deep neural networks (DNNs) has been proposed to satisfy the limited storage and computing capability of mobile edge devices. However, previous pruning methods mainly focus on reducing the model size and/or improving performance without considering the privacy of user data. To mitigate this concern, we propose a privacy-preserving-oriented pruning and mobile acceleration framework that does not require the private training dataset. At the algorithm level of the proposed framework, a systematic weight pruning technique based on the alternating direction method of multipliers (ADMM) is designed to iteratively solve the pattern-based pruning problem for each layer with randomly generated synthetic data. In addition, corresponding optimizations at the compiler level are leveraged for inference accelerations on devices. With the proposed framework, users could avoid the time-consuming pruning process for non-experts and directly benefit from compressed models. Experimental results show that the proposed framework outperforms three state-of-art end-to-end DNN frameworks, i.e., TensorFlow-Lite, TVM, and MNN, with speedup up to 4.2$\times$, 2.5$\times$, and 2.0$\times$, respectively, with 
almost no accuracy loss, while preserving data privacy. 
\end{abstract}


\maketitle

\section{Introduction}
Recent years have witnessed substantial progress and remarkable breakthroughs of deep neural networks (DNNs), especially deep convolutional neural networks (CNNs) in solving complicated visual tasks \cite{krizhevsky2012imagenet,simonyan2014very,he2016deep}. Along with the great success are the ever-increasing model size and the computing demand, which highly restrict the deployments of DNNs on mobile and edge devices with limited capacities. 
To mitigate the challenges brought by the large amount of computations and achieve the goal of real-time inference for modern DNN models, weight pruning \cite{han2015learning,dong2017learning,liu2018rethinking,zhang2018systematic, ren2019ADMMNN,wen2016learning,li2016pruning,he2018amc,Zhuang2018DCP,Zhao2019VariationalCN,zhu2018ijcai,slimming2017iccv,yang2018ISMVL,Ma2019PCONVTM} is proposed to reduce the inherent redundancy in model parameters. Early works on non-structured pruning \cite{han2015learning,dong2017learning,liu2018rethinking} prune weights at arbitrary locations using heuristic methods. Via the successful applications of the powerful Alternating Direction Methods of Multipliers (ADMM) optimization framework, later research works \cite{zhang2018systematic, ren2019ADMMNN} achieved substantial weight reduction while maintaining promising accuracy. However, non-structured pruning leads to sparse and irregular weight matrices, which require additional indices for the storage in a compact format. Consequently, these methods are not compatible with parallel hardware accelerations for the inference. By incorporating regularity into weight pruning, structured pruning \cite{wen2016learning,li2016pruning,he2018amc,Zhuang2018DCP,Zhao2019VariationalCN,zhu2018ijcai,slimming2017iccv} eliminates the requirements for weight indices, thus is more hardware friendly. On the downside, the coarse-grained nature of structured pruning degrades the accuracy more significantly. Recently, pattern-based pruning \cite{yang2018ISMVL,Ma2019PCONVTM} is proposed to inherit the benefits from fine-grained pruning while maintaining structures that can be exploited for hardware accelerations. 

Although the above-mentioned weight pruning techniques differ in sparsity schemes and pruning algorithms, most of them \cite{dong2017learning,liu2018rethinking,zhang2018systematic, ren2019ADMMNN,wen2016learning,li2016pruning,he2018amc,Zhuang2018DCP,Zhao2019VariationalCN,zhu2018ijcai,slimming2017iccv,yang2018ISMVL,Ma2019PCONVTM} are based on the assumption that the training dataset is available. However, this is not always the use case for real-world implementations. 
For example, in many areas, most notably those related to medicine, sharing data about individuals is not even permitted by law or regulations \cite{jochems2017developing,jochems2016distributed}.  Furthermore, in commercial applications, the training data should be kept as business confidentiality. 


To deal with this problem, we propose a privacy-preserving-oriented DNN pruning and mobile acceleration framework. At the algorithm level, a DNN model compression entity prunes the pre-trained models provided by users with pattern-based sparsity without the usage of any information about the private training dataset. Specifically, the pruning of the DNN model is achieved by pruning layers sequentially with randomly generated syntheic data. Instead of using the loss value, we measure the difference of the Frobenius norm between the original output of user's pre-trained model and the output of the compressed model given the same input for each layer to evaluate whether enough information is maintained after pruning. By forming the pruning problem into an optimization problem, the proposed framework solves the pattern-based pruning problems iteratively and analytically by extending the potent ADMM algorithm \cite{boyd2011distributed}. At the compiler level, corresponding pattern-enabled compiler optimizations are leveraged. After retraining the compressed model, users can achieve real-time inference without accuracy loss. The highlights of our contributions in this paper are summarized as follows: 1) We formulate the privacy-preserving-oriented pattern-based pruning problem as an optimization problem with combinatorial constraints. 2) We solve the optimization problem with an extension of the ADMM framework. 3) We accelerate the DNN execution on mobile devices with a compiler-based framework consisting of several optimizations enabled by our pattern-based design. 4) We conduct extensive experiments to compare the proposed framework with the state-of-the-art pruning methods on representative 
CNNs.

\section{Related Work}

In practice, the data used for DNN training is often massively distributed among different users, or is owned by a single party but is inconvenient or forbidden to share with others. On the one hand, users tend to store their confidential data locally for privacy concerns. On the other hand, many data owners, e.g., medical institutions, are prevented by regulations from sharing their data with others \cite{jochems2017developing,jochems2016distributed}. Meanwhile, the demand for model compression on mobile devices is imperative because of the limited capacity 
nature. 

Only few works achieve weight pruning without the original training dataset. The early weight pruning work \cite{han2015learning} proposed a magnitude-based heuristic method. Only 10$\sim$20\% of weights can be pruned without hurting accuracy when no retraining is adopted. If a higher compression rate is desired, several rounds of pruning and retraining are needed. Work \cite{Srinivas2015DatafreePP} only prunes fully-connected layers while neglecting computation intensive convolutional (CONV) layers. Recently, there are tools for data-free pruning that prunes small value weights directly, such as the one used in Cambricon AI chips. However, this approach suffers from notable accuracy loss even when removing only 20\% weights. 

To overcome the limitations of prior works and achieve real-time DNN inference, we consider to design a privacy-preserving-oriented DNN pruning and mobile acceleration framework that provides high-performance compressed model for users without the usage of the training dataset. With the proposed framework, users could directly benefit from the compressed model without privacy concerns and have no need to handle the time-consuming weight pruning process.   

\section{Framework Overview}
The overview of the proposed framework is presented in this section. We begin by introducing the threat model, then the adopted pattern-based sparsity. Next, the algorithm-level and compiler-level frameworks are demonstrated. 

\subsection{Threat Model}
We consider the following threat model in this work. Two participants, DNN compression entity and data owner, also known as the user, work together to formulate a DNN model with high performance on the private dataset of the user, as shown in Fig. \ref{fig:algo_level_opt}. The DNN compression entity has no access to the original training data but just a pre-trained DNN model from the user. This work specifically focuses on protecting the privacy of user data. We assume that the pre-trained model is trained in a trusted way and its security is out of the scope of this paper. 

\subsection{Sparse Convolution Patterns}
\begin{figure}[t!]
     \centering
     \includegraphics[width=1\columnwidth]{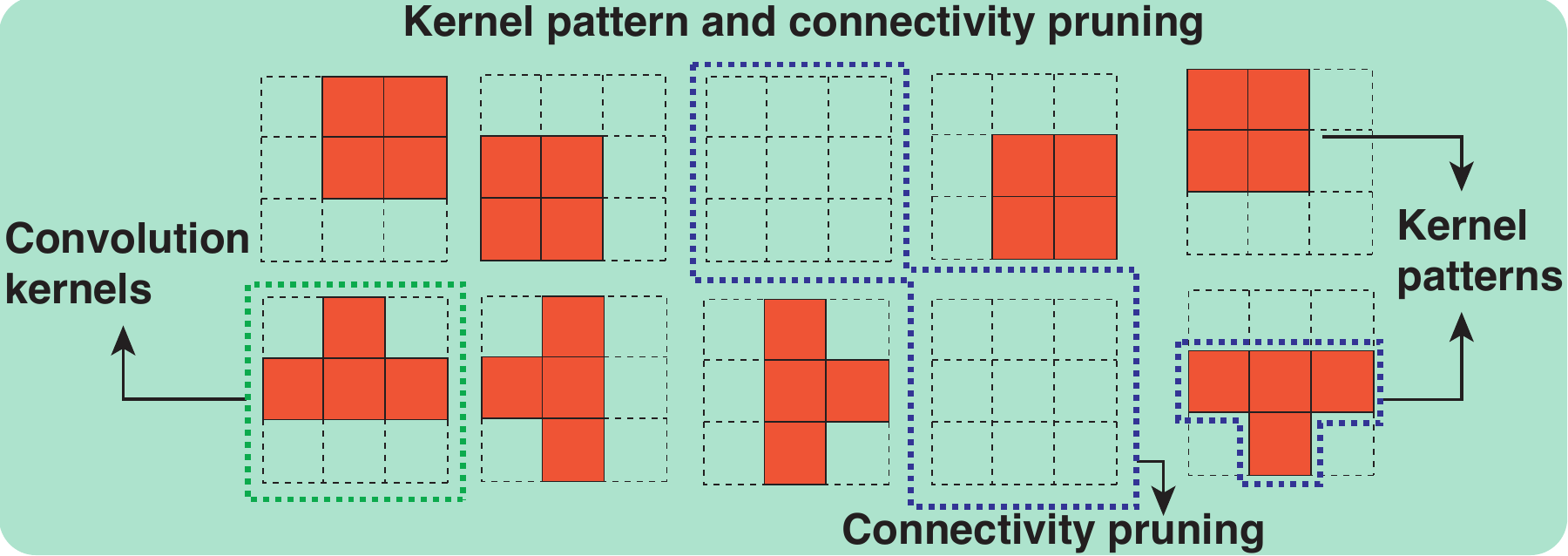}
     \caption{Pattern-based pruning.}
     \label{fig:pattern}
\end{figure}

To incorporate the advantages of both structured pruning and non-structured pruning while getting rid of the respective shortcomings, our pruning framework adopts two pattern-based pruning dimensions, i.e., kernel pattern pruning and connectivity pruning. The objective is to achieve both high inference accuracy and satisfying execution efficiency. An illustration of pattern-based pruning is given in Fig.~\ref{fig:pattern}, with orange blocks representing remaining weights while green blocks are pruned weights. 

\textbf{Kernel pattern pruning} removes weights at an intra-kernel level. The locations of the remaining weights in each kernel form a specific pattern. In this work, we focus on the kernel patterns for $3\times3$ kernels because they are widely adopted in various DNN architectures \cite{simonyan2014very,he2016deep}. Different kernels can have different patterns, but the total types of patterns are restricted to a pattern library with a fixed size. We represent the finite pattern library as $\mathcal{P} = \{\bm{M}_1,\cdots, \bm{M}_m\}$, with $m$ representing the size of the pattern library. As Fig.~\ref{fig:pattern} shows, we reserve four non-zero weights in a kernel to match the single-instruction multiple-data (SIMD) architecture of embedded CPU/GPU processors, thereby maximizing hardware throughput. 


\textbf{Connectivity Pruning} achieves inter-kernel level pruning by removing whole kernels, as illustrated in Fig. \ref{fig:pattern}. Connectivity pruning is a good supplement to kernel pattern pruning for a higher compression and acceleration rate. Both pruning schemes can be integrated into the same algorithm-level solution and compiler assisted acceleration framework. 

\subsection{Algorithm-Level Framework}
\begin{figure} [t!]
     \centering
     \includegraphics[width=0.9\linewidth]{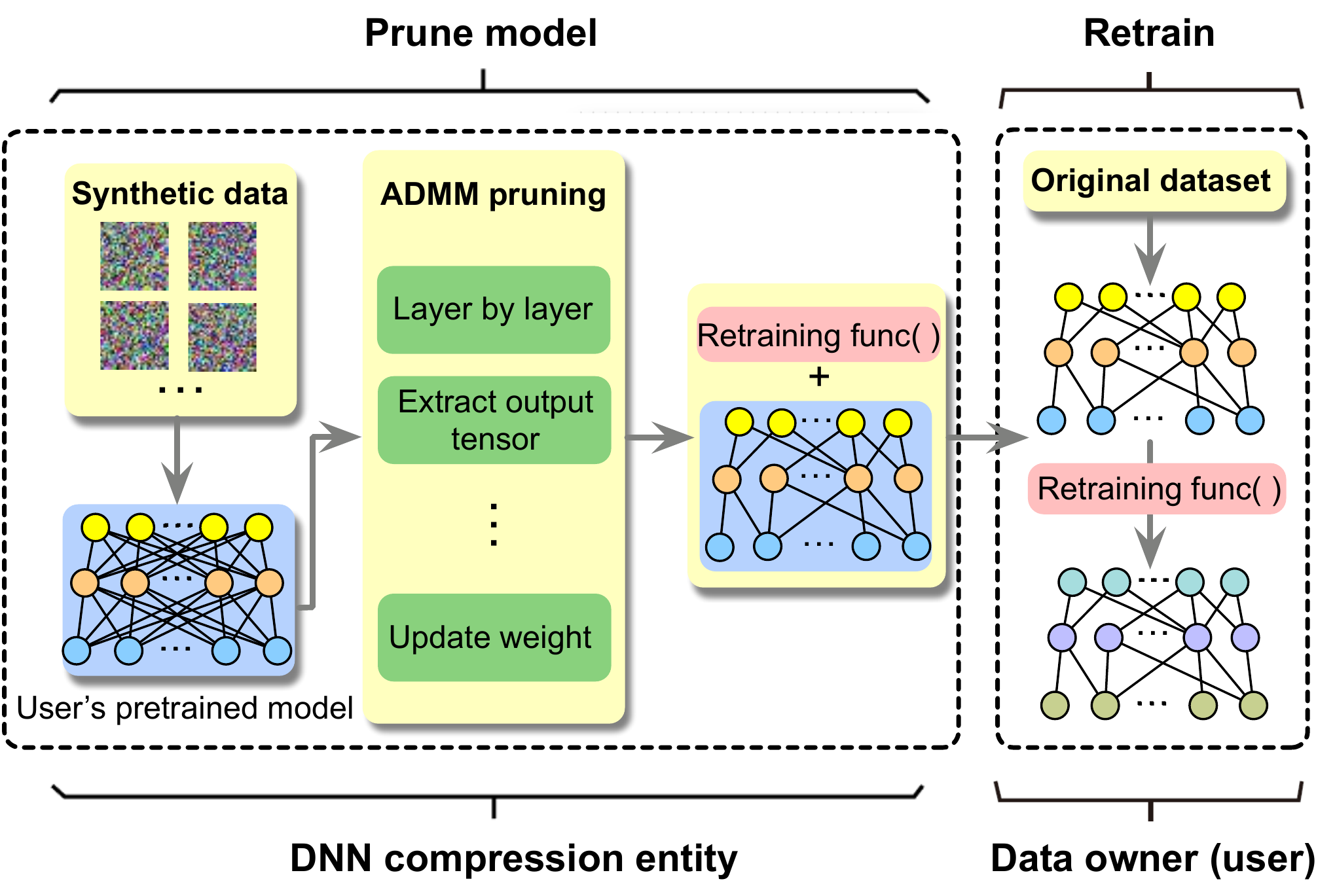}  
     \caption{{Algorithm-level framework overview.}}
     \label{fig:algo_level_opt}
\end{figure}

Fig. \ref{fig:algo_level_opt} illustrates the algorithm-level framework. The DNN compression entity is responsible for the pruning of user's pre-trained model while the user only needs to retrain the compressed model with the help of a retraining function. In the DNN compression entity side, ADMM algorithm is leveraged to achieve pattern-based pruning layer-by-layer systematically. Note that the DNN compression entity has no access to the user's training dataset but only the pre-trained model to preserve data privacy. Randomly generated synthetic data is used as the input for the pruning process. In the user side, retraining is similar to the traditional training of a DNN model, except that the retraining function sets corresponding gradients as zero for pruned weights. Only a few epochs are required before the model can make acceptable predictions. Therefore, users do not have to grasp details about pruning and can easily obtain compressed models with the proposed framework.

\subsection{Compiler-Level Framework}
\begin{figure*} [t]
     \centering
     \includegraphics[width=0.9\linewidth]{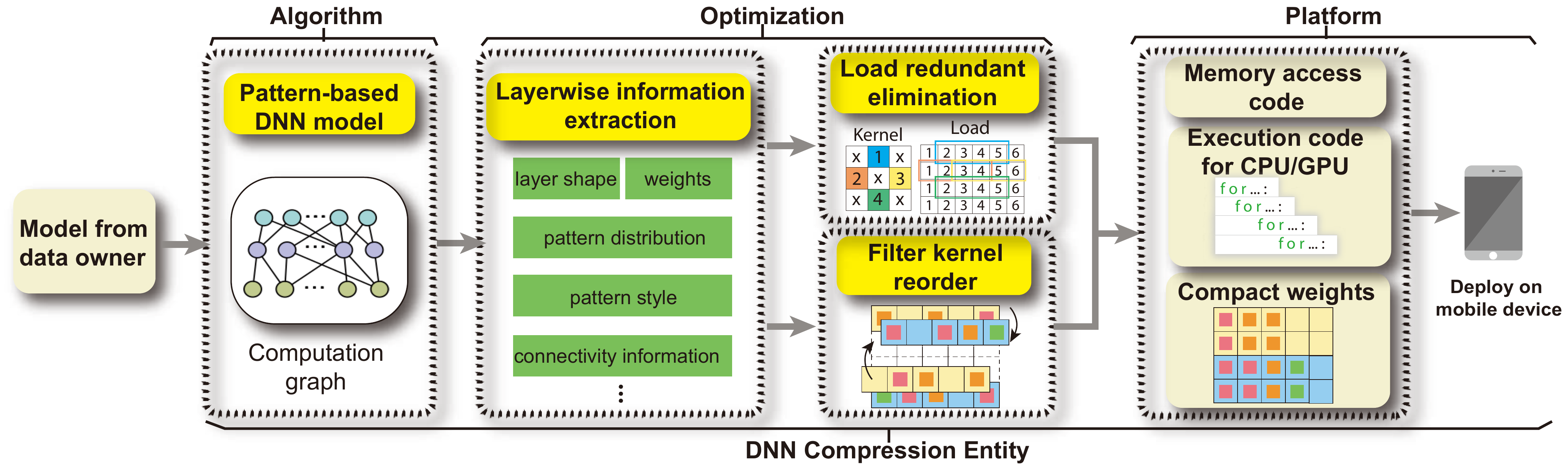}  
     \caption{{Compiler-level inference acceleration framework overview.}}
     \label{fig:pipeline}

\end{figure*}
After pattern-based pruning, we rely on a compiler-based acceleration framework to achieve real-time DNN executions on resource-restricted mobile devices, as shown in Fig. \ref{fig:pipeline}. This framework aims to address three key challenges in the pruned DNN execution: 1) heavy control-flow dependency existing within each thread; 2) computation divergence and load imbalance among different threads; 3) poor memory performance caused by irregular memory accesses.
Correspondingly, we design three pattern-enabled compiler optimizations that work on each DNN layer: filter kernel reorder, compressed weight storage, and load redundancy elimination. These optimizations are conducted on a layer-wised representation that consists of multiple parts like layer shape, pattern style, connectivity information, etc. These optimizations are general, working for both CPU and GPU code generations. 

{\bf Filter kernel reorder} addresses two challenges i) increased control-flow instructions, and ii) thread divergence and load imbalance, by grouping the filters and kernels with similar lengths/patterns together. This optimization is specifically enabled by our pattern-based pruning design.

{\bf Compressed weight storage} is specifically designed for our kernel pattern and connectivity pruning by leveraging the fact that the preserved weights follow our designed patterns. Together with filter kernel reordering, this compact data structure yields much better compression rates than the conventional CSR (compressed sparse row) format~\cite{Tinney67}.

{\bf Load redundancy elimination}~\cite{Muchnick98} addresses the poor memory performance of non-structured pruning by exploring register-level load redundancy opportunities during executable kernel code generation. 
It is crucial when data accesses between memory and cache have already been optimized using tiling~\cite{Coleman95}.


\section{Privacy-Preserving-Oriented Weight Pruning}
Our algorithm-level design is presented in this section. We first provide the formulation of the privacy-preserving-oriented pruning problem. Then the  ADMM-based solution is demonstrated. Finally, the overall pruning algorithm is outlined.
\subsection{Formulation of the Pruning Problem}
We consider the pruning of an $N$-layer DNN with a major focus on the computation-intensive CONV layers. Our objective is to find specific sparse patterns for the kernels without the usage of the original dataset. The compressed model is then sent back to the user for retraining. For each layer $n$, the weights and biases are denoted as $\bm{W}_n$ and $\bm{b}_n$, respectively. During the pruning process, the input to the DNN is $\bm{X}_0$. Note that $\bm{X}_0$ is not derived from the original training dataset, but only randomly generated synthetic data, to preserve privacy for the original training dataset. In our experiment, we set the value of each pixel within the synthetic data with a discrete uniform distribution in the range of 0 to 255. Layer $n$ takes the output $\bm{X}_{n-1}$ from the previous layer as the input volume, and produces an output volume $\bm{X}_n = \sigma(\bm{W}_n\bm{X}_{n-1} + \bm{b}_n)$, where $\sigma$ is the activation function. In order to evaluate whether representative weights are kept to maintain enough information after pruning, we measure the Frobenius norm between the original output volume $\bm{X}_n'$ of user's pre-trained model and the output $\bm{X}_n$ of the compressed model given the same input volume $\bm{X}_{n-1}$. A smaller value means that the layer can provide similar results after pruning, indicating that more information is kept. Therefore, we could formulate the problem of the pruning for the $n$-th layer as 
\begin{equation} \label{opti_ori}
\begin{split}
    &\underset{\bm{W}_n,\bm{b}_n}{\text{min}}\quad \left\lVert \sigma(\bm{W}_n\bm{X}_{n-1} + \bm{b}_n) - \bm{X}_n' \right\lVert_F^2,\\
    &\text{subject to\quad} \bm{W}_n \in C_{n}, 
\end{split}
\end{equation}
where $C_{n}$ denotes the constraint set for the $n$-th layer. $C_{n}$ restricts the pattern shapes and is defined as $C_{n}:=\{\bm{W}_n|$ each kernel in $\bm{W}_n$ needs to satisfy one specific pattern shape in the pattern set $\mathcal{P}$\}.

\subsection{Pattern Library (Set) Design}
An appropriate design of the pattern library $\mathcal{P}$ is the prerequisite to attain both good pruning results and efficient hardware implementations. The design includes the size of the pattern library and the shape of each specific candidate pattern in the library. If the size of the pattern library is too small, the pruning might not be flexible enough, thereby leading to accuracy degradation. On the contrary, it is more challenging to generate efficient codes by the compiler for hardware accelerations and finding the sparsity solution with a large pattern library size. Through empirical study, we found out that a library with $m=6-8$ patterns achieves a desirable balance between compiler overhead and accuracy for the commonly used $3\times3$ kernels.

After the size of the pattern library is settled and 4-entry patterns are utilized, the influence of patterns on the compiler and hardware is determined, regardless of specific pattern shapes. However, the pattern shapes will affect the accuracy of the compressed model and thus should be designed carefully. We select candidate pattern shapes using a simple but effective heuristic based on the following insights: 1) the central weight in a $3\times3$ kernel is critical and shall not be pruned; and 2) a smaller distortion of the kernel after pruning is preferred. Therefore, we find 3 largest weights for each kernel within the pre-trained model. The locations of the 3 largest weights together with the central weight form a 4-entry pattern. Then, $\text{top-}m$ most commonly appeared patterns in the whole DNN are selected as the candidates, forming the pattern library $\mathcal{P} = \{\bm{M}_1,\cdots,\bm{M}_m\}$, where each $\bm{M}$ only contains binary-valued elements and has the same size as the kernels. Pattern pruning is achieved by applying element-wise multiplication of $\bm{M} \in \mathcal{P}$ with kernels. A found pattern library with 8 candidate patterns is illustrated in Fig. \ref{fig:pattern}, with orange blocks representing the locations to maintain weights.  
\subsection{ADMM-based Kernel Pattern Pruning} \label{ADMMPattern}
Directly solving the optimization problem (\ref{opti_ori}) is difficult due to the non-convex constraint. Hence, we resort to the ADMM framework by decomposing the original problem into two subproblems to be solved separately. To leverage the ADMM optimization framework, we define an indicator function $I_n(\bm{W}_n)$, that is zero when the constraint  $\bm{W}_n\in C_{n}$ is satisfied, but +$\infty$ otherwise. After incorporating auxiliary variable $\bm{A}_{n}$, problem (\ref{opti_ori}) can be rewritten as
\begin{equation} \label{opti_const}
    \begin{split}
     &\underset{\bm{W}_n,\bm{b}_n}{\text{min\quad}} \left\lVert \sigma(\bm{W}_n\bm{X}_{n-1} + \bm{b}_n) - \bm{X}_n' \right\lVert_F^2 + \mathcal{I}_n(\bm{A}_{n}),\\
     & \text{subject to\quad} \bm{W}_n=\bm{A}_{n}.
    \end{split}
\end{equation}
Note that our ADMM-based method is different from previous work \cite{zhang2018systematic, ren2019ADMMNN} as we not only remove redundant weights, but also enforce certain regularity of the remaining weights in a kernel with the leverage of kernel patterns. The augmented Lagrangian \cite{boyd2011distributed} of the optimization problem (\ref{opti_const}) is given as
\begin{equation}
\begin{split}
     \mathcal{L}&(\bm{W}_n, \bm{b}_n, \bm{A}_n,\bm{D}_n ) = \left\lVert \sigma(\bm{W}_n\bm{X}_{n-1} + \bm{b}_n) - \bm{X}_n' \right\lVert_F^2\\ + &\mathcal{I}_n(\bm{A}_{n}) + \frac{\rho}{2}\left\lVert\bm{W}_n-\bm{A}_n + \bm{D}_n\right\lVert_F^2 + \frac{\rho}{2}\left\lVert \bm{D}_n \right\lVert_2^F,
\end{split}
\end{equation}
where $\bm{D}_{n}$ is the dual variable. 
To solve the problem above, we decompose it into two subproblems. At iteration $k$, the first subproblem (primal problem) is
\begin{equation} \label{subprob1}
    \underset{\bm{W}_n,\bm{b}_n}{\text{min\quad}} \left\lVert \sigma(\bm{W}_n\bm{X}_{n-1} + \bm{b}_n) - \bm{X}_n' \right\lVert_F^2 + \frac{\rho}{2}\left\lVert \bm{W}_n - \bm{A}_{n}^{k-1}+\bm{D}_{n}^{k-1}\right\lVert_F^2.
\end{equation}
Both of these two terms are differentiable and this subproblem could be solved by standard solvers such as stochastic gradient descent (SGD) effectively.  

The second subproblem (proximal problem) is given by
\begin{equation}
     \underset{\bm{A}_{n}}{\text{min\quad}}\mathcal{I}_n(\bm{A}_{n})+\frac{\rho}{2}\left\lVert \bm{W}_n^{k}-\bm{A}_{n}+\bm{D}_{n}\right\lVert_F^2. \label{subprob2}
\end{equation}
As $\mathcal{I}_n(\cdot)$ is the indicator function of the constraint set $C_n$, the globally optimal solution of the second subproblem can be derived as \cite{boyd2011distributed}
\begin{equation} 
    \bm{A}_n^{k} = \prod_{C_{n}}(\bm{W}_n^{k}+\bm{D}_{n}^{k-1}), \label{proximal_sol}
\end{equation}
where $\prod_{C_{n}}$ is the Euclidean projection of $\bm{W}_n^{k}+\bm{D}_n^{k-1}$ onto the constraint set $C_n$. The special structure of $C_n$ allows us to find the optimal analytical solutions, which is to select the pattern resulting the pruned kernel with the largest Frobenius norm, for each kernel in the layer. Then the derived $\bm{A}_n^{k}$ is fed into the primal problem in the next iteration $k+1$.

Next, we update the dual variable $\bm{D}_n^{k}$ according to
 \begin{equation}
    \bm{D}_n^{k} := \bm{D}_n^{k-1} + \bm{W}_n^{k}-\bm{A}_n^{k}.
 \end{equation}
The above alternating optimization process then proceeds to the next iteration until convergence.
\subsection{Connectivity Pruning}
Besides the above-mentioned kernel pattern pruning, we could further adopt connectivity pruning into the proposed framework to achieve a higher compression rate for users demanding a faster inference speed. Connectivity pruning can be integrated into the same algorithm-level solution in Section \ref{ADMMPattern}. We further define a constraint set $C_{n}':=\{\bm{W}_n|$ the number of nonzero kernels is no more than $\beta_n$\} for connectivity pruning, where $\beta_n$ is a predetermined hyperparameter. By replacing $\bm{C}_n$ with $\bm{C}_n'$ in the problem formulation and ADMM-based solution framework , we could obtain the results for connectivity pruning. 
\subsection{Overall Algorithm}
We formally present the overall pruning algorithm as Algorithm \ref{PruneAlg}. The DNN compression entity starts pattern-based pruning upon it receives the pre-trained model from a user. At the beginning of each iteration $k$, a batch of $T$ synthetic images are generated and preprocessed as the input for the pruning process. The pruning is going through layer-by-layer for the whole model. The generation of the synthetic images does not depend on any information about the user's private dataset. The pruning process iteratively solves the two subproblems (\ref{subprob1}) and (\ref{subprob2}) until convergence. At last, the compressed model and retraining function are released to the user for retraining. 

\begin{algorithm}[tp]
\caption{Overall Pruning Algorithm}
\label{PruneAlg}
\SetKwInOut{Input}{Input}\SetKwInOut{Output}{Output}
\Input{User's pre-trained DNN model $\{\bm{W}_n^0\}_{n=1}^N$, total iteration $K$, augmented penalty $\rho$, batch size $T$, constraint set $C_n$ for $n = 1$ to $N$}
\Output{Compressed model $\{\bm{W}_n^K\}_{n=1}^N$, the retraining function}
Initialize $ \{\bm{A}_n^0\}_{n=1}^N\leftarrow \{\bm{W}_n^0\}_{n=1}^N$, $ \{\bm{D}_n^0\}_{n=1}^N\leftarrow\bf{0}$ \;
\For{\rm{iteration} $k \leftarrow1$ \rm{to} $K$}{
Randomly generate a batch of $T$ synthetic images\;
\For{\rm{layer} $n \leftarrow1$ \rm{to} $N$}{
Get the output of the $n$-th layer from the current model and the pre-trained model\;
Update $\bm{W}_n^{k}$ by solving problem (\ref{subprob1}) with standard solvers\;
Update $\bm{A}_n^{k}$ by solving problem (\ref{subprob2}) using Eqn. (\ref{proximal_sol})\;
$\bm{D}_n^{k} := \bm{D}_n^{k-1} + \bm{W}_n^{k}-\bm{A}_n^{k}$\;
}
}
Send the pruned model $\{\bm{W}_n^K\}_{n=1}^N$ and the retraining function back to the client for retraining process\;
\end{algorithm}

\section{Experimental Results}
In this section, we present the evaluations of our privacy-preserving-oriented DNN pruning and mobile acceleration framework. We begin by measuring the performance of the algorithm-level method. Then, we demonstrate the accelerations achieved by the overall framework on mobile platforms.
\subsection{Experiment Settings}
In order to evaluate whether the proposed algorithm-level method can consistently attain efficient compressed models for tasks with different complexities, we test on three representative network structures, i.e., VGG-16, ResNet-18, and ResNet-50, with two major image classification datasets, i.e., CIFAR-10 and ImageNet. Here, CIFAR-10 and ImageNet are viewed as users' private training datasets and are not revealed to the DNN model compression entity. All these evaluations are carried out on one NVIDIA GTX 1080Ti GPU and three NVIDIA RTX 6000 GPUs. Then we experimentally analyse the execution performance of our compiler-assisted framework. 

During pruning, we use the following parameter settings for the DNN model compression entity. We initialize the penalty value $\rho= 1\times10^{-4}$, and increase $\rho$ by $10$ times for every $11$ epochs, until $\rho$ reaches $1\times10^{-1}$. SGD optimizer is utilized for the optimization steps with a learning rate of $ 1\times10^{-3}$. An epoch corresponds to 10 iterations, and each iteration process a batch of data. The batch size $T$ is set to 32. Each input image is generated by setting the value of each pixel with a discrete uniform distribution in the range of 0 to 255.

To demonstrate the acceleration of pattern-based sparsity provided by our framework on mobile devices, we compare the proposed framework with three state-of-the-art DNN inference acceleration frameworks, i.e., TFLite \cite{TensorFlow-Lite-Results}, TVM \cite{chen2018tvm}, and MNN \cite{Ali-MNN}. Our experiments are conducted on a Samsung Galaxy S10 cell phone with the latest Qualcomm Snapdragon 855 mobile platform that consists of a Qualcomm Kryo 485 Octacore CPU and a Qualcomm Adreno 640 GPU.


\begin{table}[] 
\caption{\textbf{Comparison results on CIFAR-10 dataset}}
\setlength{\tabcolsep}{0.8mm}{ 
\label{table1}\scalebox{0.84}{
\begin{tabular}{|c|ccccc|}
\hline
 & Methods & \begin{tabular}[c]{@{}c@{}}Base \\ Accuracy\end{tabular} & \begin{tabular}[c]{@{}c@{}}Prune\\ Accuracy\end{tabular} & \begin{tabular}[c]{@{}c@{}}Conv\\ Comp.Rate\end{tabular} & \begin{tabular}[c]{@{}c@{}}Sparsity\\ (Pattern)\\ Type\end{tabular} \\ \hline
 \multirow{7}*{\rotatebox{90}{ResNet-18}} & 
 DCP\cite{Zhuang2018DCP} & 88.9\% & 87.6\% & 2.0$\times$ & Structured \\
 & AMC\cite{he2018amc} & 90.5\% & 90.2\% & 2.0$\times$ & Structured \\
 & Variational Pruning\cite{Zhao2019VariationalCN} & 92.0\% & 91.7\% & 1.6$\times$ & Structured \\
 & \textbf{Privacy-Preserving} & 94.1\% & 94.9\% & 8$\times$ & Pattern \\
 & \textbf{Privacy-Preserving} & 94.1\% & 94.5\% & 12$\times$ & Pattern \\
 & \textbf{Privacy-Preserving} & 94.1\% & 94.2\% & 16$\times$ & Pattern \\ \hline
 \multirow{5}*{\rotatebox{90}{ResNet-50}} & 
 One Shot Pruning\cite{liu2018rethinking} & 93.8\% & 93.6\% & 2.5$\times$ & Irregular \\
 & AMC\cite{he2018amc} & 93.5\% & 93.5\% & 1.7$\times$ & Structured \\
 & \textbf{Privacy-Preserving} & 94.2\% & 95.0\% & 8$\times$ & Pattern \\
 & \textbf{Privacy-Preserving} & 94.2\% & 94.7\% & 12$\times$ & Pattern \\
 & \textbf{Privacy-Preserving} & 94.2\% & 94.4\% & 16$\times$ & Pattern \\ \hline
 \multirow{8}*{\rotatebox{90}{VGG-16}} & 
 Iterative Pruning\cite{han2015learning}\cite{liu2018rethinking} & 92.5\% & 92.2\% & 2.0$\times$ & Irregular \\
 & One Shot Pruning\cite{liu2018rethinking} & 92.5\% & 92.4\% & 2.5$\times$ & Irregular \\
 & 2PFPCE\cite{min20182pfpce} & 92.9\% & 92.8\% & 4.0$\times$ & Structured \\
 & Efficient ConvNet \cite{li2016pruning} & 93.2\% & 93.4\% & 2.7$\times$ & Structured \\
 & \textbf{Privacy-Preserving} & 93.5\% & 93.1\% & 8$\times$ & Pattern \\
 & \textbf{Privacy-Preserving} & 93.5\% & 92.4\% & 12$\times$ & Pattern \\
 & \textbf{Privacy-Preserving} & 93.5\% & 91.6\% & 16$\times$ & Pattern \\ \hline
\end{tabular}}
}
\end{table}

\subsection{Accuracy and Compression Rate Evaluations}
We first experiment on CIFAR-10 dataset with the VGG-16, ResNet-18, and ResNet-50 networks. As shown in Table \ref{table1}, our method not only preserves data privacy, but also reaches a very high compression rate, which is comparable or even better than other pruning algorithms that use the original training dataset during pruning. For ResNet-18, we achieve a 16$\times$ compression rate and 94.2\% accuracy after pruning. And for ResNet-50, we achieve a 16$\times$ compression rate and 94.4\% accuracy after pruning. Our method also works well on VGG-16, with a 16$\times$ compression rate and 91.6\% accuracy.

With promising results on CIFAR-10, we also investigate the performance of our method on ImageNet using ResNet-18. We achieve a 4$\times$ compression rate with almost no top-5 accuracy degradation on ResNet-18, which is much better than Network Slimming and DCP. We could further reach a 6$\times$ compression rate with 88.0\% top-5 accuracy. Moreover, it only takes 14 hours for the proposed method to finish the pruning process using one RTX 6000 GPU.

Furthermore, we notice that our proposed method can even achieve accuracy improvement compared to the pre-trained model on various network structures. The accuracy improvement is attributed to the enhanced image processing ability with the leverage of patterns. Note that performing pattern pruning solely can reserve a 2.25$\times$ compression rate due to the intra-kernel sparsity.
\begin{table}[]
\caption{\textbf{Comparison results on ImageNet dataset}}
\setlength{\tabcolsep}{0.8mm}{
\scalebox{0.84}{
\begin{tabular}{|c|ccccc|}
\hline
 & Methods & \begin{tabular}[c]{@{}c@{}}Base \\Top-1/5\\ Accuracy\end{tabular} & \begin{tabular}[c]{@{}c@{}}Prune\\Top-1/5\\ Accuracy\end{tabular} & \begin{tabular}[c]{@{}c@{}}Conv\\ Comp.Rate\end{tabular} & \begin{tabular}[c]{@{}c@{}}Sparsity\\ (Pattern)\\ Type\end{tabular} \\ \hline
 \multirow{4}*{\rotatebox{90}{ResNet-18}} & 
 Network Slimming\cite{slimming2017iccv} & 68.9/88.7\% & 67.2/87.4\% & 1.4$\times$ & Structured \\
 & DCP\cite{Zhuang2018DCP} & 69.6/88.9\% & 69.2/88.8\% & 3.3$\times$ & Structured \\
 & \textbf{Privacy-Preserving} & 69.9/89.1\% & 69.3/89.0\% & 4$\times$ & Pattern \\
 & \textbf{Privacy-Preserving} & 69.9/89.1\% & 68.0/88.0\% & 6$\times$ & Pattern \\ \hline
\end{tabular}}}
\end{table}

\subsection{Performance Evaluation on Mobile Platform}
In this part, we demonstrate our evaluation results on a mobile device to show the real-time inference of our proposed pattern-based sparse model with the help of the compiler-based acceleration framework. To guarantee fairness, the same pattern-based sparse model are used for all frameworks, and the fully optimized configurations of TFLite, TVM and MNN are enabled.

\begin{figure} [t!]
     \centering
     \includegraphics[width=0.9\linewidth]{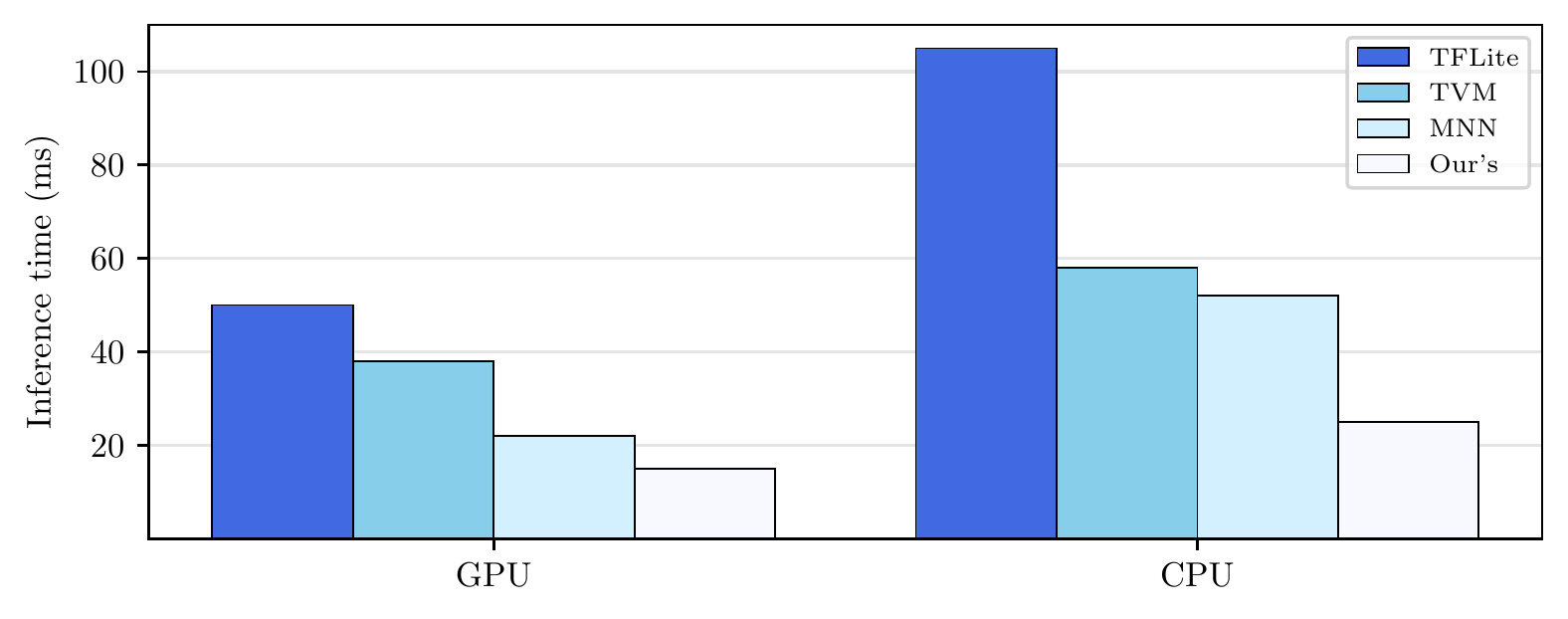}  
           \vspace{-0.4 cm}
     \caption{{Compiler-based acceleration on 6$\times$ compressed ResNet-18 with ImageNet.}}
     \label{mobile}
     
\end{figure}

Fig. \ref{mobile} shows the mobile CPU/GPU execution time of the pattern-based model on different platforms. The testing model is ResNet-18 with a 6$\times$ compression rate on ImageNet dataset. We can observe that our approach achieves significant acceleration on mobile devices compared with other frameworks. On CPU, the proposed framework achieves 4.2$\times$ speedup over TFLite, 2.3$\times$ speedup over TVM, and 2.0$\times$ speedup over MNN. On GPU, our framework achieves 3.3$\times$ speedup over TFLite, 2.5$\times$ speedup over TVM and 1.4$\times$ speedup over MNN. This is because previous frameworks such as TFLite, TVM, and MNN do not have specific optimizations for compressed models as leveraged in our framework. Therefore, there is no obvious acceleration in real implementations on mobiles with such frameworks even though the models have been highly compressed. Real-time execution typically requires 30 frames/sec, i.e., 33ms/frame. From our results, all of our DNN models meet or far exceed this requirement. Furthermore, some of them can even accomplish real-time inference on a mobile CPU.  

\section{Conclusion}
In this paper, we propose a privacy-preserving-oriented DNN pruning and mobile acceleration framework. At the algorithm level, we formulate the problem of pattern-based pruning without the usage of the original training dataset as an optimization problem and solve it successfully with an extension of the powerful ADMM. At the compiler level, we adopt corresponding pattern-enabled optimizations. Extensive experiments demonstrate that the proposed framework can achieve real-time inference and maintain accuracy on representative large-scale CNNs while preserving data privacy. 

\section{Acknowledgement}
This work is partly supported by the National Science Foundation (CCF-1901378, CCF-1919117, CCF-1937500, and CNS-1909172).

\bibliographystyle{IEEEtran}  
\bibliography{sample-base}
\end{document}